# Forecast-Then-Optimize Deep Learning Methods

**Jinhang Jiang** [a]
jinhang.jiang@mckesson.com

**Nan Wu** [a]
nan.wu@mckesson.com

**Ben Liu** [b]
ben.liu@walmart.com

**Mei Feng** [c]
meifeng@ku.edu

**Xin Ji** [a]
jane.ji@mckesson.com

**Karthik Srinivasan** [d]
karthiks@ku.edu

a. Enterprise Forecasting, McKesson Corp.
b. International Operation, Walmart Inc.
c. Biopharmaceutical Innovation & Optimization Center, The University of Kansas
d. School of Business, The University of Kansas

## Abstract

Time series forecasting underpins vital decision-making across various sectors, yet raw predictions from sophisticated models often harbor systematic errors and biases. We examine the Forecast-Then-Optimize (FTO) framework, pioneering its systematic synopsis. Unlike conventional Predict-Then-Optimize (PTO) methods, FTO explicitly refines forecasts through optimization techniques such as ensemble methods, meta-learners, and uncertainty adjustments. Furthermore, deep learning and large language models have established superiority over traditional parametric forecasting models for most enterprise applications. This paper surveys significant advancements from 2016 to 2025, analyzing mainstream deep learning FTO architectures. Focusing on real-world applications in operations management, we demonstrate FTO's crucial role in enhancing predictive accuracy, robustness, and decision efficacy. Our study establishes foundational guidelines for future forecasting methodologies, bridging theory and operational practicality.

## 1 Introduction

Time flows like a river, carrying with it patterns of the past that ripple into the future. The art and science of forecasting attempt to decipher these patterns, transforming uncertainty into informed decisions. Time series forecasting is a cornerstone of data-driven decision-making which underpins critical applications in operations management. Forecasting aims to predict future values of a temporal sequence based on historical observations, enabling proactive planning and risk mitigation (Qi et al. 2022, Y. Wang et al. 2025). Forecasting plays a crucial role across various organizational functions, from strategic planning to daily operational control. It underpins inventory management, production scheduling, capacity planning, and supply chain coordination, ultimately impacting an organization's efficiency, profitability, and customer satisfaction (Choi et al. 2018, Nguyen 2023). The increasing complexity of modern business environments, characterized by volatile markets, intricate supply chains, and rapidly evolving consumer demands, necessitates the adoption of sophisticated forecasting techniques that can handle vast amounts of data and capture intricate patterns (Hanifan and Timmermans 2018). Historically, time series forecasting has evolved from statistical models like AutoRegressive Integrated Moving Average (ARIMA) (Box 2013), Exponential Smoothing (ETS) (Brown 1959), and Gaussian Processes (Roberts et al. 2013) to machine learning approaches such as Random Forests (Breiman 2001) and Gradient Boosting Machines (GBM) (Friedman 2001). These methods remain strong contenders in many forecasting scenarios, offering robustness, interpretability, and reliability, particularly when data is limited or highly structured. However, as forecasting problems grow in scale and complexity, capturing intricate temporal dependencies and nonlinear patterns becomes increasingly challenging for these traditional approaches. (Y. Wang et al. 2025).

Meanwhile, deep learning has become a powerful tool for capturing complex relationships and adapting to diverse forecasting tasks (Ismail Fawaz et al. 2019, Torres et al. 2020, Lim and Zohren 2021, Wen et al. 2022, Jin, Koh, et al. 2023, Liu and Wang 2024, Y. Wang et al. 2025). A key advancement is the use of deep learning as global forecasting models (GFMs) (Januschowski et al. 2020, Hewamalage et al. 2022, Bandara 2023). Such (global) deep learning models leverage cross-learning across multiple time series to capture shared patterns to improve efficiency and generalization, especially in large-scale forecasting.

Building on these advancements in forecasting for operations management (OM), a critical next step is to ensure that forecast outputs are as accurate, reliable, adaptive and robust as possible before they are used for managerial decision-making. This necessity has led to the emergence of the Forecast-then-Optimize (FTO) framework, which explicitly separates the forecasting phase from a subsequent post-processing or refinement phase. FTO eextends the predict-then-optimize (PTO) framework for decision-making (Elmachtoub and Grigas 2022), where forecasts serve as inputs to an optimization model that determines the best course of action, such as inventory management or pricing strategies. In contrast, FTO shifts the focus to optimizing the forecasts themselves and acknowledges that raw model outputs often contain biases, systematic errors, or suboptimal patterns that require further adjustments.

At its core, FTO consists of two primary components: (1) a forecasting phase, where models generate predictions based on historical data, and (2) an optimization (post-processing) phase, where various techniques—such as residual adjustments, ensemble methods, meta-learners, and model selection—are applied to refine these forecasts. The optimization stage does not alter the fundamental structure of the forecasting model(s) but rather enhances its outputs by diluting accumulated error over time, recalibrating uncertainties, or incorporating multiple model perspectives to improve accuracy (Ganaie et al. 2021, Wu and Levinson 2021, Mohammed and Kora 2023).

In the context of enterprise forecasting, the FTO framework bridges the gap between raw model predictions and practically useful forecasts, ensuring that outputs are not only statistically sound but also better aligned with real-world conditions and downstream applications. This additional layer of enhancement is particularly crucial in high-stakes environments, where even marginal forecast improvements can lead to significant operational and financial benefits (Jiang et al. 2025).

This review examines advancements in the FTO framework in the past decade (2016–2025), focusing on deep time series models, post-processing, ensemble, model selection, and meta-learning techniques to improve forecasting accuracy. We define the FTO framework, highlight the need for forecast adjustments, and compare optimization approaches. Key deep learning models including recurrent neural networks (RNNs), convolutional neural networks (CNNs), transformers, multi-layer perceptrons (MLPs), and graph neural networks (GNNs), are reviewed in the context of forecasting, alongside enhancements like standardization, decomposition, and Fourier methods. We have also explored applications in supply chain, healthcare, energy, and finance before concluding with challenges in scalability and deployment, and future directions in optimization and uncertainty-aware post-processing.

## 2 Forecasting in Operations Management

Several highly cited OM papers have addressed various aspects of forecasting and its impact on operational decisions. Kahneman and Lovallo (1993) explored the cognitive perspective on risk-taking and the role of bold forecasts in managerial decision-making. (Tam and Kiang 1992) demonstrated the managerial applications of neural networks in the context of bank failure predictions. Chen et al. (2000) quantified the bullwhip effect in supply chains, highlighting the significant impact of forecasting, lead times, and information sharing on inventory variability. Elmachtoub and Grigas (2022) introduced the Smart "Predict, then Optimize" (SPO) framework, underscoring the critical link between prediction and optimization in effective forecasting for decision-making. A foundational review of the field was provided by Makridakis and Wheelwright in their book (Makridakis, Spyros G., Wheelwright 1979), offering a comprehensive overview of available forecasting methods. Then, (Zhao et al. 2002) examined the impact of forecasting model selection on the value of information sharing within a supply chain, emphasizing the practical implications of forecasting accuracy. Makridakis and Hibon (2000) publication on the M3-Competition presented the results and conclusions of a major forecasting competition, serving as a benchmark for evaluating different forecasting methods. (Fildes and Goodwin 2008) review provided a comprehensive overview of forecasting and its intersection with operational research, highlighting key developments and opportunities in the field. One study (Petropoulos et al. 2014) emphasized the critical importance of selecting appropriate forecasting methods based on the characteristics of the demand data. (Banerjee et al. 2020) review focused on passenger demand forecasting in scheduled transportation, showcasing the application of forecasting in a specific operational context. (Goltsos et al. 2022) review provides an in-depth analysis of inventory forecasting, a core component of operations management. A review of guidelines for the use of combined forecasts offered practical insights into improving forecasting accuracy by leveraging multiple forecasts (De Menezes et al. 2000). Research on the judgmental selection of forecasting models (Petropoulos et al. 2018) highlights the enduring role of human expertise in the forecasting process. Research on analytics for multiperiod risk-averse newsvendor problems under nonstationary demands, integrates forecasting with decision-making under uncertainty, a crucial aspect of operations (Y. Yu et al. 2023). Demand forecasting with supply-chain information and machine learning methods has shown promise in the pharmaceutical industry (Zhu et al. 2021), highlighting the value of information sharing and advanced techniques. Another paper explored the integration of forecasting with inventory decisions, focusing on profit optimization (Lin et al. 2022).

Table 1 provides a summary of a few key forecasting studies in operations management.

**Table 1: Selected Forecasting Studies in Operations Management**

| Title of Paper | Citation | Key Contribution/Focus | Journal Name |
|---|---|---|---|
| Timid choices and bold forecasts—A cognitive perspective on risk-taking | (Kahneman and Lovallo 1993) | Explores the psychological aspects of forecasting and decision-making. | Management Science |
| Reducing the cost of demand uncertainty through accurate response to early sales | (Fisher and Raman 1996) | Links forecasting accuracy to the reduction of costs associated with demand uncertainty. | Journal of Operations Management |
| Quantifying the bullwhip effect in a simple supply chain: The impact of forecasting... | (Chen et al. 2000) | Quantifies the impact of forecasting accuracy on supply chain dynamics. | Management Science |
| The M3-Competition: results, conclusions and implications | (Makridakis and Hibon 2000) | Presents the findings of a major forecasting competition, benchmarking various methods. | European Journal of Operational Research |
| Optimising forecasting models for inventory planning | (Kourentzes et al. 2020) | Focuses on optimizing forecasting models specifically for inventory management purposes. | International Journal of Production Economics |
| Demand Forecasting with Supply-Chain Information and Machine Learning... | (Zhu et al. 2021) | Highlights the use of advanced techniques and information sharing for improved demand forecasting. | Production and Operations Management |

| Forecasting gold price with the XGBoost algorithm and SHAP interaction values | (Jabeur et al. 2024) | Showcases the application of advanced machine learning for forecasting in finance. | Annals of Operations Research |
| --- | --- | --- | --- |
| The effect of visibility on forecast and inventory management performance during the COVID-19 pandemic | (Dehkhoda et al. 2025) | Examines the role of forecasting in managing inventory during a major disruption. | International Journal of Production Research |
| Forecasting Urban Traffic States with Sparse Data Using Hankel Temporal Matrix Factorization | (Chen et al. 2024) | Presents a sophisticated computational method for forecasting in transportation. | INFORMS Journal on Computing |

## 3 Forecast-Then-Optimize (FTO) Methodology

### 3.1 Definition

Forecasting and optimization are two interrelated processes. Forecasting comprises methods used to predict future data points before their actual observation. On the other hand, optimization seeks to improve a particular process or model to achieve the best possible outcome, often concerning cost, efficiency, or accuracy based on specific criteria. Forecast-Then-Optimize (FTO) is a two-phase methodology framework designed to enhance forecasting accuracy through model-driven prediction followed by systematic forecast adjustment. The first phase, forecasting, constructs multiple predictive models from historical data to capture temporal dependencies, seasonality, and nonlinear interactions. This phase generates baseline forecasts that serve as the foundation for further refinements. The second phase, optimization or post-processing, refines these forecasts generated from the first phase to improve accuracy and generalizability. Unlike traditional approaches that rely solely on initial predictions, FTO acknowledges that no single model can capture all scenarios, and the likelihood of identifying the true model in large-scale systems is low. By explicitly addressing biases through historical error structures and complementary model strengths, the framework proactively enhances forecast reliability and robustness while reducing the need for extensive tuning or refinement in the initial modeling phase. Figure 1 is a high-level illustration of the FTO framework.

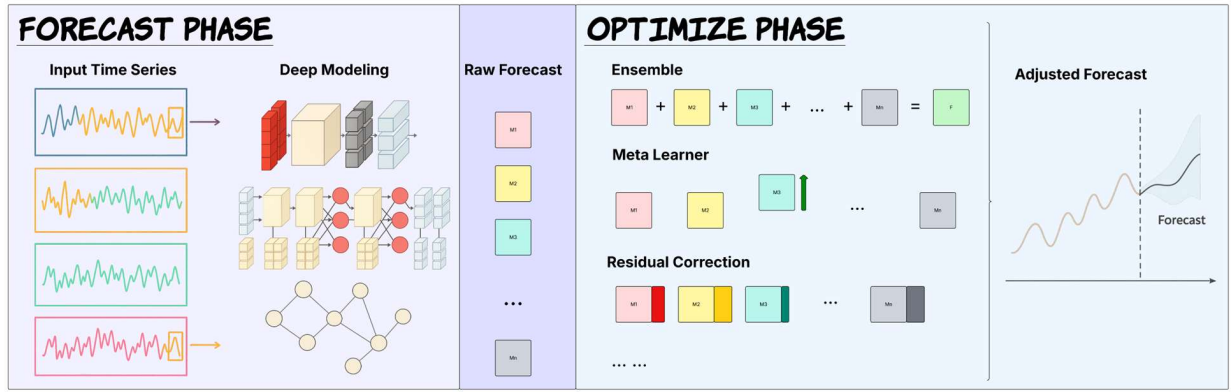

Figure 1. An illustration of the FTO Framework

### 3.2 Rationale

Purely model-driven forecasts, even those utilizing advanced architectures like Transformers, often encounter challenges in effectively capturing complex temporal dependencies and inherent patterns in time series data. Recent studies have highlighted that simpler linear models can outperform these sophisticated models, particularly in long-term forecasting scenarios, suggesting that increased model complexity does not always equate to improved performance (Zeng et al. 2023). Additionally, the computational intensity of large language models (LLMs) applied to time series forecasting has not resulted in corresponding performance improvements without additional processing or specialized transformation, further questioning the efficacy of complex architectures in this domain (Tan et al. 2024).

While advanced models strive to capture complex temporal dependencies, there are inherent limits to what can be achieved purely through modeling. The optimization phase provides an opportunity to squeeze additional value from forecasts, supported by findings from the M4 Competition, which demonstrated that combining multiple forecasting techniques often outperforms individual models, highlighting the potential of adjustments to enhance forecast reliability (Makridakis et al. 2018).

### 3.3 Current FTO Methods

Within the FTO framework, the guiding assumption is: "No matter how good the forecasts are, once they are finalized after training, there is always room to squeeze more accuracy." This principle reflects the value of optimizing existing forecasts rather than relying exclusively on improvements to the underlying models. While hyperparameter tuning can enhance the performance of global deep learning models, its impact is rather limited when compared to post-processing optimization strategies. Unlike the computationally expensive search for optimal parameters, post-processing optimization techniques, such as bias correction, meta-learning, and ensemble methods, offer a more efficient way to refine forecasts, bringing improved accuracy with lower resource overhead, especially in situations where users prefer to freeze the original algorithms in their legacy systems.

We consider a scenario where a pool of candidate models has been developed for a large-scale time series dataset, and the challenge is to adjust, combine, or rank them to maximize global accuracy and forecasting benefits. While K-fold cross-validation is commonly used for model tuning and selection, sliding-window-based methods like Prequential in Sliding Blocks (Preq-Sld-Bls) are more prevalent in time series forecasting (Cerqueira et al. 2021), as they dynamically adjust models by evaluating their recent performance to better capture temporal patterns. Traditional statistical criteria such as Akaike information criterion (AIC) and Bayesian information criterion (BIC) have also been used to assess model performance but are often impractical for large-scale deep learning models due to resource constraints or limited applicability. Therefore, we focus on handy, scalable and robust alternatives to optimize model selection and combination. Based on method characteristics, we categorize these approaches into two groups based on their characteristics: Ensemble methods and Meta-learners.

### 3.3.1 Ensemble Methods

Ensemble techniques improve predictions by averaging the predictions of the candidate models, in hopes of benefiting from their complementary strengths. Static weight ensembles assign a constant weight to each prediction model and compute weighted average as refined predictions. It is easy to implement and interpret and also highly scalable; even equally weighted average ensembles have long been shown to be more accurate than single models (Makridakis et al. 2018). Among static weight ensembles, averaging (mean) and median ensembles are the most commonly used in practice. The logic for mean ensembles is to average predictions across models assuming that under squared-error loss, the mean minimizes expected error if component forecasts are unbiased, and their individual errors will cancel each other out, resulting in a more stable and better prediction (X. Wang et al. 2022). This is particularly helpful when models are diverse but reasonably good, as it reduces variance without significantly increasing bias. If all models are similar or suffer a common bias, even an ensemble will not improve much (Thomson et al. 2019). Recent deep-learning ensemble studies (Ganaie et al. 2021) also affirm that combining heterogeneous model architectures yields more robust predictions.

On the other hand, the core principle behind median ensembles is to construct a stable aggregation method that remains resilient to outliers and erratic predictions (X. Wang et al. 2022). By computing the median forecast at each time step, this approach effectively suppresses the influence of unstable or low-performing models, making it particularly well-suited for high-variance environments. In large-scale real-world applications, median ensembles are often favored due to their robustness against extreme values and minimal tuning complexity, offering a pragmatic balance between accuracy and simplicity. Beyond the standard median ensemble, several extensions have been proposed. One common generalization is quantile averaging, where instead of computing the 0.5 quantile (the median), forecasts are aggregated using other quantiles such as the 0.1, 0.25, or 0.9 quantiles to match specific risk preferences or forecast objectives. This method provides asymmetric aggregation behavior, which is particularly useful in contexts such as inventory planning or energy forecasting, where under- and over-predictions incur different costs. Quantile averaging can thus be useful for probabilistic forecasting and risk-sensitive decisions, ensuring that forecasts align with a target service level or risk threshold. For example, combining forecasts via the median (which is the 50% quantile) has been used to stabilize COVID-19 predictions (Ray et al. 2022). Another extension is the weighted median ensembles, which attempt to further improve performance by assigning higher weights to more accurate or more relevant models. Research in econometrics shows that giving weights inverse to past errors or based on models' relative performance can yield gains (X. Wang et al. 2022). The weighted median is defined as the point in the sorted set of forecasts where the cumulative weight reaches 0.5, combining the outlier resistance of the median with the adaptivity of performance-based weighting. Together, these quantile- and weight-based strategies provide interpretable, robust ensemble tools particularly suitable for large-scale and high-stakes forecasting applications. Static weights, however, are inherently less proactive as they do not respond to changes in model performance or data shifts. When the reliability of the candidate models varies over time, which is common in real-world forecasting, static ensembles continue to apply fixed assumptions, thus resulting in suboptimal performance. Moreover, they also face another critical limitation: when the pool of candidate models includes significant weak learners, the inherited errors from these models persist within the ensemble, ultimately pulling the aggregated forecasts away from the true demand (X. Wang et al. 2022).

Dynamic weighting ensembles (DWE) and dynamic ensemble selection (DES) are closely related strategies for adaptive model combination, both designed to improve forecast accuracy by responding to changes in model performance over time. The primary difference of those two methods lies in their aggregation behaviors. DWE assigns weights to all candidate models, generating a continuous combination of forecasts, while DES selects a subset of models, sometimes even a single model, based on contextual or recent performance criteria. In practice, DES can behave like a DWE when it assigns nonzero weights to multiple models, making the two approaches functionally similar under certain conditions. Both methods can incorporate techniques such as sliding window evaluations, machine learning, or reinforcement learning to update model selection or weighting based on incoming data (Jin et al. 2022, Zhao et al. 2024). For instance, one recent study proposed a reinforcement learning-based ensemble that treats model combination as a sequential decision-making process, learning to adjust weights in response to evolving time series dynamics (Fu et al. 2022). These dynamic approaches often achieve higher forecasting accuracy than static ensembles by adapting to regime shifts and changing model performance (Galicia et al. 2019), although they typically require more computational resources. A related method, Bayesian model averaging (BMA) (Leamer and Leamer 1978), provides a probabilistic framework that combines forecasts by estimating the posterior probability of each model given the observed data, enabling effective forecasting adjustments in various domains (Zhang and Yang 2015, Barigou et al. 2023). BMA produces calibrated, uncertainty-aware forecasts and has demonstrated strong performance across a range of applications. By adjusting a regularization parameter, often denoted as lambda, BMA can flexibly assign weights to all candidate models, a subset, or even a single model, depending on the underlying uncertainty and data fit. When applied in an adaptive or recomputed fashion, BMA can be considered a principled variant of DES, offering both the interpretability of Bayesian inference and the adaptability of dynamic model selection.

Moreover, hierarchical forecasting, widely adopted in supply chain domain (Babai et al. 2022), addresses structured time series that follow aggregation constraints, aiming to ensure coherence across levels. While base forecasts are generated independently at each level, reconciliation techniques adjust them to respect hierarchical relationships. Common methods include bottom-up, top-down, middle-out, and MinT (Hyndman and Athanasopoulos 2021, Babai et al. 2022), with MinT offering state-of-the-art accuracy by leveraging forecast error covariances (Wickramasuriya and Hyndman 2020). In the context of the FTO framework, reconciliation acts as a post-processing layer and often integrates with ensemble or selection strategies applied at each node. Recent advancements have explored neural reconciliation and learning-based adjustments, further aligning hierarchical forecasting with modern ensemble modeling practices (Wickramasuriya and Hyndman 2020, Olivares et al. 2022).

### 3.3.2 Meta-learners

In this section, we focus on two types of meta-learners: bias correction-based and model selection-based. Like ensemble methods, they are also developed to provide simple yet model-agnostic and domain-agnostic ways to improve forecasting accuracy over the candidate models.

A simple residual bias correction method adjusts the forecasts by averaging the recent forecasts errors or other heuristics. Such residual adjustments are easy to implement and scale by just adding a bias term, but they cannot fix complex patterns of errors. A straightforward machine learning-based approach is residual learning, where a secondary model is trained to predict the difference between actual outcomes and the initial forecast, known as Direct Loss Estimator (DLE) (NannyML (release 0.13.0) 2023). The forecast is then adjusted by this predicted residual. This technique is model-agnostic as it treats the initial model as a black box and has proven effective in practice (Yin and Liu 2022, Tedesco et al. 2023). In parallel to these meta-learning strategies, uncertainty-aware methods adjust and augment forecasts to account for risk and confidence. One widely used approach is quantile forecasting (Wen et al. 2018, Rodrigues and Pereira 2020), unlike quantile averaging whose objective is to find a single data point from a pool of forecasters, the model predicts quantiles of the future outcome distribution. This provides decision-makers with a range of plausible outcomes and their likelihood. Quantile regression methods are increasingly used to capture tail risks and variability in fields like finance and supply chain (Mitchell et al. 2023). Another emerging approach, conformal forecasting (Alaa and van der Schaar 2023, Xu and Xie 2023), complements quantile forecasting by providing empirically calibrated prediction intervals that carry finite-sample probabilistic guarantees, enhancing confidence in forecast reliability. Building on these methods, practitioners often implement ad-hoc adjustments using quantiles to offset systematic over- or underforecasting. Nevertheless, such uncertainty-aware strategies tend to be less adaptive compared to model-based adjustments and can suffer from hysteresis effects.

Besides training models to learn and fix residuals, another route is to treat forecast model selection as a learning problem itself, aiming to choose the best candidate model(s) for each scenario or time series (Barak et al. 2019, Y. Li et al. 2021). A general framework involves three steps: training a pool of candidate models, labeling the data based on a selection criterion, and then training a meta-learner to predict the best model. Concept-wise, this approach is very similar to MoE where a gating layer is used to select the most relevant candidate model for the problem. When multiple candidate models are selected and combined together, it functions as a subvariant of DES. FFORMA (Montero-Manso et al. 2020) extends this idea using gradient-boosted trees trained on extracted features to generate weights for candidate models. Unlike static ensembles, FFORMA offers instance-specific model weighting, enabling strong performance on short and heterogeneous series as evidenced by its top performance in the M4 competition. Classification-Based Model Selection (CMS) is one such method that has been applied in retail demand forecasting (Ulrich et al. 2022). The researchers demonstrated that this automated model selection can outperform any single forecasting method across a heterogeneous product set, improving overall profitability by matching each demand pattern to its most suitable model (Ulrich et al. 2022). An emerging trend frames model selection as a language modeling task, inspired by the analogy between choosing the next word in a sentence and choosing the best model for a time series. TimeSpeaks adopts this view by tokenizing time series data and using a Transformer-based architecture to learn the mapping between sequences and their most suitable forecasting models. This generative approach over a vocabulary of models enables the system to capture structural and temporal cues in a language-like fashion, offering strong generalization in large-scale or heterogeneous forecasting settings, particularly for long-horizon tasks (Jiang et al. 2025).

Another lightweight alternative is Forecast Selection and Representativeness (REP) (Petropoulos and Siemsen 2023), which mimics human judgment by comparing out-of-sample forecasts to historical patterns, even before future values are observed. Unlike traditional selection criteria like AIC or cross-validation, REP introduces an asynchronous evaluation that penalizes forecasts that deviate from learned patterns. By combining a performance gap and representativeness gap, REP outperformed statistical baselines across more than 100,000 time series from M1–M4 competitions (Petropoulos and Siemsen 2023). Though not explicitly trained as a meta-learner, REP serves as a low-cost, instance-based selection method, offering a practical alternative in large-scale and diverse forecasting environments.

Compared to ensemble methods, selecting a single best model simplifies the problem and avoids aggregating potentially conflicting signals. It can also be effective in large-scale applications where each candidate model specializes in a certain time series. But it risks overfitting and can be unstable when multiple models perform similarly. Ensemble methods are generally safer and more robust over time, though they may introduce additional noise if poorly constructed and require careful weighting to avoid performance dilution.

## 4   Deep Forecasting Models

In this section, we will review the state-of-the-art models based on deep learning. First, we will provide an overview of the models in subsection 4.1, and dive into their details in the following sub-subsections. We will discuss the trade-offs of different types of models in subsection 4.2. Lastly, we will discuss the emerging techniques and tricks in the deep time series algorithm development in subsection 4.3. Figure 2 illustrates a timeline of influential forecasting models developed over the past decade. These models were selected based on their groundbreaking contributions to the field, either by pioneering novel architectures or techniques, or by achieving widespread adoption in industry. A clear evolution is evident: the early years saw the adoption of RNNs, CNNs and GNNs, which laid the foundation for deep time series modeling. This was followed by a shift toward MLPs and Transformers, which introduced more scalable and expressive architectures. Most recently, LLM-based models have gained traction in research, signaling a new era of general-purpose forecasting approaches.

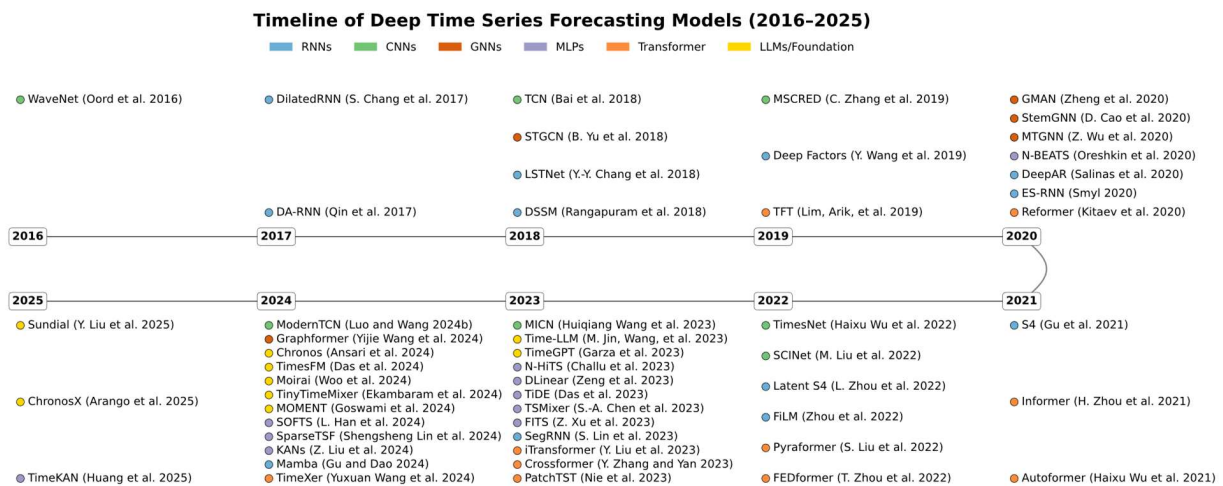

Figure 2. The Timeline of Select Deep Time Series Algorithms

## 4.1 Overview of Deep Time Series Models

The selection of models discussed within this paper was guided by rigorous criteria emphasizing recent advancements, performance on standardized benchmarks, architectural novelty, and practicality to large-scale forecasting problems. The chosen models span significant breakthroughs in deep learning-based forecasting from 2016 to 2025, guaranteeing a survey of up-to-date methodologies. We surveyed forecasting studies in top conferences and journals, including NeurIPS, ICML, AAAI, IJCAI, SIGKDD, IEEE TPAMI, Management Science, European Journal of Operations Research, International Journal of Forecasting, Journal of Operations Management, Production and Operations Management, to shortlist the list of models. Furthermore, models such as DeepNPTS, N-BEATS, N-HiTS, DLinear, PatchTST, SOFTS, TiDE, TimeMixer, and TSMixer, have demonstrated a range of architectural paradigms, ranging from entirely linear models that challenge the necessity for complexity, to sophisticated hybrid solutions featuring multi-scale decomposition and hierarchical forecasting. The models were also evaluated in terms of practical applicability, scalability, and robustness, particularly emphasizing their ability to work under real-world conditions and competitions like the M4 and M5 forecasting challenges.

### 4.1.1 State-Based Sequential Models

Recurrent Neural Networks (RNN)s and their gated variants, such as Long-short term memory (LSTM) and Gated Recurrent Unity (GRU), were among the earliest deep learning models applied to time series forecasting . These models maintain hidden states over time, making them well-suited for one-step and multi-step predictions. DeepAR (Salinas et al. 2020), an early attempt to train a global RNN across multiple time series, improved forecast accuracy by learning shared patterns and outputting probabilistic predictions. Models like DilatedRNN (Chang et al. 2017) extended RNNs to longer sequences, while DA-RNN (Qin et al. 2017), an encoder-decoder LSTM with dual attention, mitigated noise and long-sequence challenges, demonstrating strong performance in economic and energy forecasting. Hybrid architectures were also explored, such as LSTNet (Chang et al. 2018), which combined a 1-D CNN to capture short-term patterns with an LSTM for long-term dependencies, stabilizing training and improving long-horizon accuracy. Researchers further sought to integrate classical forecasting principles with RNNs. BRITS (Cao et al. 2018) introduced a data-driven imputation strategy, enhancing forecast reliability, while Deep State Space Models (DSSM) (Rangapuram et al. 2018), Structured State Space Sequence Models (S4) (Gu et al. 2021), and Latent S4 (LS4) (L. Zhou et al. 2022) aimed to bridge state-space modeling (SSMs) (Hamilton 1994) with deep learning. The emergence of state-based sequential models, encompassing both RNNs and state-space models, has provided a broader framework for modeling time-dependent data. While RNNs rely on sequential recurrence, SSMs leverage structured state evolution to process temporal dependencies efficiently. Deep Factors (Wang et al. 2019) blended global deep learning with local probabilistic models, and Recurrent Neural Filters (Lim, Zohren, et al. 2019) incorporated Bayesian filtering into RNNs for probabilistic forecasting. The ES-RNN (Smyl 2020), which fused ETS with LSTMs, famously won the M4 competition by a wide margin. As RNNs exhibited limitations for long-sequence forecasting, Frequency Improved Legendre Memory Model (FiLM) employed Legendre Polynomial Projections to efficiently store historical information, avoiding the gradient vanishing/explosion issues of RNNs. It further incorporates a Fourier-based filtering mechanism to remove noise from time series data, ensuring that critical patterns are preserved while avoiding overfitting. By combining these components with a low-rank approximation for computational efficiency, FiLM outperformed transformer-based models and other deep learning approaches (T. Zhou, Ma and Wang Qingsong Wen Liang Sun Tao Yao Wotao Yin Rong Jin 2022). SSMs have gained traction in time series forecasting because they combine the memory advantages of RNNs with the efficiency of matrix-based updates. Unlike RNNs, which struggle with $O(L)$ computational complexity, SSMs achieve $O(\log(L))$ or better, making them highly scalable. *Mamba* (Gu and Dao 2024), the successor to SSMs, builds on these advancements by introducing a structured state-space approach tailored for long-range forecasting. Instead of sequential state updates, Mamba selectively retrieves past information using parallel state-space transformations, significantly improving efficiency while preserving temporal dependencies. Recent research has empirically validated Compared to transformer-based models (reviewed in 3.1.5), Mamba achieves a strong balance between performance and computational efficiency, reducing memory usage and training time while maintaining superior forecast accuracy (Z. Wang et al. 2025). The Simple-Mamba (S-Mamba) model demonstrates that Mamba can effectively capture both inter-variate correlations and temporal dependencies through its bidirectional encoding layer, outperforming leading Transformer-based models on multiple real-world datasets. While it excels in periodic and multivariate datasets, its effectiveness on aperiodic and low-variability datasets is limited due to weaker cross-variate dependencies. Additionally, models like SegRNN (Lin et al. 2023) introduced segment-wise iterations and parallel multi-step forecasting to improve long-term performance. Furthermore, the Water-wave Information Transmission and Recurrent Acceleration Network (WITRAN) framework captures both long- and short-term repetitive patterns through bi-granular information transmission while modeling global and local correlations using the Horizontal Vertical Gated Selective Unit (HVGSU) (Jia et al. 2023). However, RNN-based models like WITRAN still suffer from efficiency issues in computational performance. Thus, TPGN, a new successor to RNNs, introduces the Parallel Gated Network (PGN), which reduces the information propagation path by extracting and fusing historical information with gated mechanisms. To further enhance long-range forecasting, TPGN employs two branches: PGN for long-term periodic patterns and patch-based representations for short-term information aggregation (Jia et al. 2024).

RNN-based models marked the beginning of deep learning for time series, significantly improving on classical baselines by capturing complex temporal dynamics. The evolution toward state-based sequential models broadened the range of approaches, with SSMs and Mamba emerging as a scalable alternative to RNNs for long-range forecasting. Hybrid approaches and attention mechanisms addressed many of RNNs' weaknesses, paving the way for modern architectures like Transformers (Lim, Arik, et al. 2019). While newer models dominate long-range forecasting, RNNs remain valuable for moderate sequence lengths and continue to serve as core building blocks in more advanced forecasting frameworks. Open-source implementations of many of these models have facilitated widespread adoption in both research and industry (Alexandrov et al. 2020, Olivares et al. 2022).

### 4.1.2 Convolution-Based Temporal Models

Though originally for vision, **CNNs** proved effective for time series by capturing local patterns with translation-invariant filters (Casolaro et al. 2023). 1-D CNNs can parallelize sequence processing and avoid recurrent computations, making them faster and often easier to train than RNNs (Bai et al. 2018).

A pioneering deep CNN for sequence modeling, WaveNet (Oord et al. 2016, Wu et al. 2019), was introduced for raw audio generation, leveraging dilated convolutions to expand the receptive field, enabling the capture of long-range dependencies without recurrence. This concept was later applied to time series forecasting, demonstrating CNNs' ability to model complex temporal patterns efficiently. SCINet (M. Liu et al. 2022), a recursive downsample-convolve-interact architecture, further advanced CNN-based forecasting by extracting multi-resolution temporal features, improving long-term dependencies through hierarchical feature aggregation. Beyond direct sequence modeling, MSCRED (Zhang et al. 2019) introduced a novel approach by transforming time series into structured images, using CNNs to detect subtle anomalies in multivariate data.

Empirical studies have also shown that generic CNN architectures can outperform RNNs on sequence tasks, leading to the development of the Temporal Convolutional Network (TCN) (Bai et al. 2018), followed by refinements like MICN (Wang et al. 2023) and ModernTCN (Luo and Wang 2024b). Moreover, TimesNet (Wu et al. 2022) redefined CNN-based forecasting by converting 1-D time series into a 2-D representation, introducing TimesBlock as a general-purpose backbone to capture multi-periodic patterns adaptively, further demonstrating CNNs' versatility in time series modeling.

CNN-based models introduced speed and long-horizon capability to time series forecasting. Dilated convolutions (Oord et al. 2016, Bai et al. 2018) enabled modeling dependencies over hundreds of time steps while avoiding the trainability issues of RNNs. CNNs have also shown strong performance in classification and anomaly detection, where local pattern recognition is crucial. However, as this review focuses on forecasting, we will not explore these applications in detail. By 2020, hybrid models combining CNNs and RNNs became common (Livieris et al. 2020, Rostamian and O'Hara 2022, Alshingiti et al. 2023), leveraging fast parallel convolutions for short-term feature extraction and sequential layers for capturing long-term dependencies and causal structure.

### 4.1.3 Graph-Based Relational Models

*GNNs* emerged to handle spatiotemporal data or inter-dependent multiple time series, where the internal interaction of data, feature and/or model can be represented by a computational graph of nodes and edges. For example, in traffic forecasting, the road network was represented by a graph and each traffic node will proceed a time series data from sensor. In this scenario, GNNs were used to capture relationships among nodes where temporal model, such as CNNs or RNNs, was created for individual time series data of each node.

Spatio-Temporal Graph Convolutional Networks (STGCN) were among the first to apply graph convolutions for spatial dependencies and 1-D convolutions for temporal dependencies in traffic forecasting, offering faster training and fewer parameters while capturing complex spatiotemporal patterns (Yu et al. 2018). The Diffusion Convolutional RNN (DCRNN) advanced this by modeling traffic flow as a diffusion process on a directed graph, integrating random walk-based diffusion convolution within an RNN to enhance long-term forecast stability (Li et al. 2018). Building on this, DGCRN introduced dynamic graph learning, allowing the graph structure to evolve over time (Li et al. 2023). Multivariate Time-series GNN (MTGNN) extended graph-based modeling beyond physical networks by learning K-nearest-neighbor graphs in feature space, enabling its application to any multivariate time series, such as financial data (Wu et al. 2020). Graph Multi-Attention Network (GMAN) utilized spatiotemporal attention, applying attention to both graph structure and temporal positions, achieving high forecasting accuracy under dynamic conditions but at a higher computational cost (Zheng et al. 2020). Spectral Temporal Graph Neural Network (StemGNN) combined spectral graph convolution with Fourier analysis, enhancing multivariate forecasting by capturing periodic patterns (Cao et al. 2020). Most recently, Graphformer replaced the self-attention mechanism in Transformers with a graph self-attention mechanism, automatically learning implicit sparse graph structures to improve long-sequence forecasting accuracy (Yijie Wang et al. 2024). More recently, T-PATCHGNN (Zhang et al. 2024) addresses the unique challenges of irregular multivariate time series by combining patch-level temporal modeling with time-adaptive graph neural networks. Each univariate series is transformed into variable-length patches with unified time resolution, avoiding sequence explosion from alignment. Inter-variable relationships are captured using dynamically learned graphs applied at the patch level, allowing the model to handle both irregularity and asynchrony efficiently. T-PATCHGNN achieves strong forecasting performance across diverse scientific domains while remaining computationally scalable.

Graph-based models have significantly advanced forecasting in domains with inherent network structures, such as traffic, transportation, energy grids, and epidemiology. They effectively capture complex joint patterns, such as rerouting effects in traffic, that traditional models struggle with. Beyond physical networks, the ability to learn graphs from data (Wu et al. 2020) has enabled these models to uncover dependencies in high-dimensional time series, such as sensor arrays and economic indicators, where relationships are not fully known.

### 4.1.4 Feedforward Time Series Models

MLPs, also known as feed-forward networks, seem too simple for sequence data, but were shown success at time series forecasting from recent architectural studies. By stacking fully connected layers with clever architectural design, these models are able to capture trends and seasonality and are very fast to train, compared to other models. They often operate on "windows" of the time series as input rather than one step at a time.

N-BEATS (Oreshkin et al. 2020) pioneered this approach, introducing a deep MLP architecture with backward and forward residual links to decompose time series into trend and seasonality components. N-HiTS (Challu et al. 2023) extended this by introducing multi-scale hierarchical interpolation, improving long-horizon forecasting through signal decomposition at different resolutions. DLinear and NLinear (Zeng et al. 2023) challenged the need for complex architectures, showing that simple window-based linear models can rival deep models in long-horizon forecasting. Around the same time, TiDE (Das et al. 2023) introduced a hybrid MLP model integrating temporal and dynamic embeddings to better adapt to complex time series structures. TSMixer (Chen et al. 2023) proposed a novel approach by modeling both time-mixing and feature-mixing patterns, capturing dependencies across temporal and variate dimensions. FITS (Xu et al. 2023) demonstrated that a highly compact architecture with only 10k parameters can still achieve competitive forecasting accuracy. Meanwhile, several models explored alternative perspectives. DeepNPTS (Rangapuram et al. 2023) moved away from parametric assumptions, generating forecasts by sampling from the empirical distribution, allowing it to capture non-Gaussian uncertainties. FreTS (Yi et al. 2023) introduced frequency-domain MLPs, leveraging spectral learning for improved energy compaction and forecasting accuracy. Koopa (Liu, Li, et al. 2023) applied Koopman theory, disentangling time-variant and time-invariant components for robust non-stationary forecasting. FTMLP (Haoxin Wang, Mo, et al. 2024) introduced feature-temporal blocks within MLPs to better model multivariate dependencies. More recently, SOFTS (L. Han et al. 2024) incorporated series-specific adaptations to improve generalization, while TimeMixer (Shiyu Wang, Wu, et al. 2024) utilized multiscale decomposition for hierarchical forecasting. WPMixer (Murad et al. 2024) further advanced this line of work by combining multi-level wavelet decomposition with patch and embedding mixers. By processing wavelet coefficients through resolution-specific branches, it captures both local and global patterns across time and frequency domains, outperforming both MLP- and Transformer-based models in long-horizon tasks with significantly improved efficiency. Building on the trend of extreme lightweight modeling, SparseTSF (Lin et al. 2024)pushes the parameter efficiency frontier further by proposing a Cross-Period Sparse Forecasting technique. It downsamples sequences based on known periodicity, performs sparse sliding prediction using a single shared linear layer, and reconstructs the forecast via upsampling. Despite using fewer than 1,000 parameters, SparseTSF matches or outperforms many SOTA models across long-horizon benchmarks. Its design is well-suited for resource-constrained settings and demonstrates strong generalization, especially on periodic datasets, making it a compelling entry in the family of channel-independent, MLP-style models. PatchMLP (Tang and Zhang 2024) adopted a patch-based mechanism, enhancing sequence locality to better capture long-term temporal patterns. These advancements underscore the continuous innovation in MLP-based forecasting.

Kolmogorov–Arnold Networks (KANs) (Z. Liu et al. 2024) are a novel neural network architecture inspired by the Kolmogorov–Arnold representation theorem, which states that any multivariate continuous function can be decomposed into a finite composition of univariate functions and additions. Recognized as a potential successor to MLPs, KANs have been explored for time series modeling due to their strong interpretability and adaptability. Recent studies show that KANs effectively model time series data, particularly in multivariate forecasting. The Multi-layer Mixture-of-KAN (MMK) network introduces a Mixture-of-KAN (MoK) layer, utilizing a mixture-of-experts (MoE) approach to adaptively assign different KAN variants to specific variables, improving forecasting accuracy while preserving interpretability (X. Han et al. 2024). Additionally, T-KAN and MT-KAN extend KANs' capabilities, with T-KAN detecting concept drift and modeling dynamic dependencies, while MT-KAN enhances multivariate forecasting by learning cross-variable interactions (Xu et al. 2024). Most recently, TimeKAN (Huang et al. 2025)rethinks forecasting as a frequency decomposition task. It employs a Decomposition–Learning–Mixing framework that cascades moving average filters, applies multi-order KANs to learn frequency-specific patterns, and recombines the signals for final prediction. Despite its lightweight design, TimeKAN achieves outstanding performance in long-horizon forecasting benchmarks, outperforming many Transformer-based models with significantly lower computational cost (Huang et al. 2025).

MLP-based models have proven that complex sequence handling like sequence or convolution is not strictly necessary for accurate forecasting. The resurgence of MLPs influenced new research into GFMs (Januschowski et al. 2020). MLP-based time series models brought a combination of accuracy, speed, and simplicity, and reminded the community that sometimes a well-designed fully-connected network can rival more sophisticated architectures.

### 4.1.5     Transformer-Based Sequence Models

***Transformers*** revolutionized NLP with self-attention mechanisms (Vaswani et al. 2017) and were adapted for time series to address long-range dependency issues inherent in RNNs. Transformers process sequences in parallel and use attention to focus on relevant time steps, making them appealing for long-horizon forecasting and complex seasonal patterns.

LogTrans (Li et al. 2019) introduced log-sparse attention, restricting queries to a logarithmic number of keys, proving that Transformers could outperform RNNs in long-horizon forecasting. TFT (Lim, Arik, et al. 2019) combined self-attention with gating mechanisms to integrate static features, covariates, and historical data, achieving strong interpretability in multi-horizon forecasting. Reformer (Kitaev et al. 2020) and Linformer (Wang et al. 2020) improved computational efficiency, using locality-sensitive hashing and low-rank projections, respectively, though domain-specific models ultimately outperformed them in forecasting tasks. Autoformer (Wu et al. 2021) pioneered seasonal-trend decomposition within transformer architectures, replacing standard self-attention with Auto-Correlation, which leveraged periodic relationships to enhance efficiency and accuracy in long-horizon forecasting. Pyraformer (S. Liu et al. 2022) introduced pyramidal attention, organizing time points hierarchically to efficiently model short- and long-term dependencies, influencing later models like Informer (Zhou et al. 2021), which used ProbSparse Self-Attention to reduce quadratic complexity and self-attention distillation for long-sequence forecasting. FEDformer (T. Zhou, Ma, Wen, et al. 2022) further enhanced efficiency by incorporating Fourier-based attention, assuming sparse frequency spectra in time series, while ETSformer (Woo et al. 2022) integrated ETS into Transformers to better model level and seasonal components. Non-stationary Transformers (Y. Liu et al. 2022) address over-stationarization by combining Series Stationarization for predictability and De-stationary Attention to retain intrinsic non-stationary patterns, enhancing adaptability in real-world forecasting. Recent innovations have focused on refining representation learning and scalability for multivariate forecasting. iTransformer (Liu, Hu, et al. 2023) restructured the Transformer by inverting input dimensions, embedding time points as variate tokens to enhance multivariate correlation modeling. Crossformer (Zhang and Yan 2023) introduced Two-Stage Attention (TSA), attending separately to time and variable dimensions, capturing complex dependencies in high-dimensional data like traffic and weather. DSformer (C. Yu et al. 2023) introduced Double Sampling (DS) to extract global and local features, combined with Temporal Variable Attention (TVA) to model dependencies across multiple timescales. BasisFormer (Ni et al. 2023) enhances time series forecasting by learning interpretable bases through self-supervised contrastive learning, computing similarity coefficients via bidirectional cross-attention, and refining predictions with a Forecast module. PatchTST (Nie et al. 2023) applied patch-based tokenization, segmenting time series into patches and using channel-independent attention to prevent certain variables from dominating the learning process, improving efficiency and long-horizon accuracy. TimeXer (Yuxuan Wang et al. 2024) advanced exogenous variable integration, applying patch-wise self-attention for temporal modeling and variate-wise cross-attention to reconcile external influences, effectively bridging causal relationships between exogenous and endogenous variables. Further refinements have targeted periodicity and adaptability. Peri-midFormer (Wu et al. 2024) modeled multi-periodic variations using a periodic pyramid structure, capturing inclusion and overlap relationships across timescales while applying self-attention to extract temporal dependencies. Ada-MSHyper (Shang et al. 2024) introduced adaptive hypergraph learning, leveraging group-wise interactions and multi-scale pattern extraction to improve forecasting accuracy. Finally, DeformableTST (Luo and Wang 2024a) tackled the over-reliance on patching, replacing it with deformable sparse attention to independently identify important time points, eliminating the need for long input sequences and large patch sizes while maintaining efficiency.

The Transformer revolution has profoundly shaped 2020s research, with a growing number of top-conference papers advancing time series each year. Beyond academic benchmarks, these models are now being tested in real-world applications such as supply chain demand prediction. Their continued refinement enhances efficiency, scalability, and adaptability, making them increasingly robust for diverse time series forecasting challenges.

### 4.1.6 General Purpose Time Series Models

In recent years, the development of foundation models for time series forecasting has gained significant momentum, leading to the emergence of several notable architectures. This section provides a concise overview of prominent general-purpose time series models, highlighting their unique features and contributions to the field.

Chronos (Amazon) (Ansari et al. 2024) applies tokenization via scaling and quantization, allowing transformer-based encoder-decoder architectures to perform zero-shot probabilistic forecasting across various benchmarks. In addition to Chronos, Amazon has released sub-variants such as Chronos-Bolt, which is up to 250 times faster and 20 times more memory-efficient than the original Chronos models of the same size, and ChronosX (Arango et al. 2025), which is specifically designed to better incorporate exogenous features into LLM-based foundation models. Time-LLM (Alibaba Group) (Jin, Wang, et al. 2023) reprograms off-the-shelf language models through modified input representations and fine-tuning strategies. TimesFM (Google) (Das et al. 2024), a decoder-only transformer, introduces patching, where windows of time points serve as tokens, enabling efficient modeling of local temporal structures. Pretrained on 100 billion real-world time points, it has demonstrated strong zero-shot forecasting performance. Alongside these adaptations, several models are natively designed as foundation models for time series forecasting. TimeGPT (Nixtla) (Garza et al. 2023) is among the first of its kind, offering superior accuracy and inference speed compared to Chronos and TimesFM. ForecastPFN (Abacus.AI) (Dooley et al. 2023), a zero-shot forecasting model that was trained exclusively on synthetic data, using a prior-data fitted network (PFN) designed to approximate Bayesian inference. Through extensive experiments, ForecastPFN demonstrates superior accuracy and efficiency compared to approaches that utilize significantly more training data. Sundial (Tsinghua University) (Liu et al. 2025) introduces TimeFlow Loss, a flow-matching approach that eliminates the need for discrete tokenization, enabling pretraining on continuous-valued time series. Pretrained on 1 trillion time points, Sundial (Y. Liu et al. 2025) achieves state-of-the-art zero-shot forecasting. Moirai (Salesforce) (Woo et al. 2024), built on a masked encoder-based transformer, is trained on LOTSA (27 billion observations across nine domains) and excels at zero-shot forecasting of trends and seasonal patterns. Its enhanced version, Moirai-MoE, integrates a MoE transformer, allowing for token-level model specialization, leading to further performance gains. Meanwhile, efficiency-focused models have been developed to balance performance and computational constraints. TinyTimeMixer (IBM) (Ekambaram et al. 2024) is a lightweight time series model optimized for resource-constrained environments, ensuring fast inference without sacrificing accuracy. MOMENT (Carnegie Mellon University) (Goswami et al. 2024) takes a different approach by addressing the lack of public time series repositories. Trained on Time-series Pile, a large and diverse collection of public datasets, MOMENT is designed as a general-purpose model capable of handling forecasting, classification, anomaly detection, and imputation, even with limited supervision. Moreover, Timer-XL (Tsinghua University) (Y. Liu et al. 2024) is a causal Transformer for unified time series forecasting, extending next-token prediction to multivariate forecasting. Its decoder-only architecture models causal dependencies across varying-length contexts, while TimeAttention captures fine-grained intra- and inter-series dependencies using flattened time series tokens (patches).

LLM-based time series modeling is an exciting development at the intersection of NLP and forecasting. It essentially asks: can we treat time series like a "language" and leverage the emergent capabilities of LLMs? Early evidence says yes, to an extent that is scientifically intriguing. Recent empirical studies evaluated LLM-based forecasting vs. tuned traditional models. It found that while LLMs, like GPT-3.5, can handle many cases, they sometimes fall short on very volatile or highly nonstationary data, where specialized models or transformations are needed. The conclusion was that LLMs are not a silver bullet for all time series yet, but they offer a powerful new baseline, especially when quick forecasts are needed without model training (Tan et al. 2024, Tang et al. 2025).

### 4.1.7 Summary of Deep Learning in Forecasting

Over the past decade, deep learning has transformed time series forecasting, evolving from basic autoregressive models to architectures capable of capturing multi-year patterns, complex dependencies, and providing interpretability. Each model category has contributed uniquely: RNNs introduced sequence modeling, CNNs improved efficiency, GNNs incorporated relational awareness, MLPs offered speed and simplicity, Transformers enabled long-range dependencies, and LLMs introduced generalization and few-shot learning. Rather than one approach dominating, the best results often come from hybrid models that combine their strengths, such as using Transformers for global patterns, GNNs for spatial relationships, and MLPs for fine-tuning biases. A key theme in this evolution has been the balance between inductive biases and data-driven learning. Early models relied on explicit trend and seasonality modeling, while flexible architectures like N-BEATS and Transformers later learned these patterns automatically, yet structured enhancements like Autoformer and FEDformer still provided performance gains. The rise of foundation models such as GPT-based approaches signals a future where pretrained generalist models may complement or even replace specialized forecasting models in some contexts. As deep learning continues to advance, the interplay between pure learning and guided learning remains crucial, shaping the next generation of time series forecasting (Gunasekaran et al. 2024, Qi et al. 2025).

## 4.2 Trade-offs among Various Architectures

Time series forecasting encompasses a diverse range of deep learning architectures, including RNNs, CNNs, GNNs, Transformers, and LLMs, each offering distinct trade-offs in accuracy, efficiency, interpretability, and practicality. Selecting the right model requires balancing technical performance with business considerations such as cost, scalability, and deployment feasibility. Below, we examine these architectures, their optimal use cases, and their impact across real-world applications, including supply chain, finance, healthcare, and energy forecasting.

### 4.2.1 RNNs

RNNs have long been a cornerstone of deep learning for sequential data, excelling at capturing short- and medium-term temporal dependencies. They are often a practical first choice when working with moderate data sizes and forecasting tasks that do not require extreme long-range memory (Gu et al. 2021). LSTMs and GRUs have been widely adopted in applications like finance and retail, offering improved gating mechanisms to control information flow (Cao et al. 2019, Falatouri et al. 2022).

One major drawback of vanilla RNNs is their sequential computation, where each step depends on the previous one, limiting parallelization and making training slower on long sequences. However, their modest parameter count ensures a manageable memory footprint and computational efficiency, allowing them to be deployed on single GPUs or even CPUs in resource-constrained environments. Newer variants like xLSTM introduce architectural tweaks that improve scalability, achieving performance comparable to Transformers while avoiding the quadratic scaling of attention mechanisms (Beck et al. 2024).

For long-horizon forecasting, RNNs often struggle with memory limitations (Gu et al. 2021), requiring strategies like sequence-to-sequence architectures or scheduled sampling to mitigate error accumulation. While they can capture fine-grained patterns, they may underperform when required to model seasonal trends or long-range dependencies (e.g., annual energy consumption cycles). Also, RNNs are largely black-box models, limiting interpretability. Enhancements like attention mechanisms can help identify salient time steps and features, as seen in ICU patient mortality prediction, where attention weights aligned with known clinical risk factors (Gandin et al. 2021).

Overall, RNNs remain reliable and efficient for moderate-length forecasting tasks, particularly when computational resources are limited. They integrate seamlessly with structured covariates (Fiterau et al. 2017) and remain a pragmatic choice for real-world deployments where cost, scalability, and interpretability must be balanced. While they may not always lead in accuracy benchmarks, their robustness and practicality ensure their continued relevance in time series forecasting.

### 4.2.2 CNNs

CNNs have proven effective in time series forecasting by treating sequences as 1-D signals. Unlike RNNs, which process data sequentially, CNNs compute filter responses across all time steps simultaneously, allowing TCNs (Bai et al. 2018) and other CNN-based architectures to apply sliding filters over time steps for efficient local pattern detection. Their key strength is pattern recognition, making them particularly useful in domains with multi-scale temporal structures, such as energy forecasting, where daily cycles are embedded within weekly or seasonal trends. Some CNN-based models have even rivaled Transformers in forecasting, classification, and anomaly detection benchmarks, provided their receptive field is well-configured (Y. Wang et al. 2025).

CNNs also offer fast training and inference, which can leverage optimized matrix operations that run efficiently on both GPUs and CPUs. Compared to Transformers, CNNs require fewer parameters, reducing memory consumption and overfitting risks on smaller datasets. In real-time forecasting applications, such as high-frequency trading (Chen et al. 2018, Peng et al. 2024) or smart grid control (Hasan et al. 2019), CNNs can process thousands of time steps within milliseconds, significantly outpacing RNNs while maintaining accuracy. They also perform well in embedded and IoT applications for image tasks (Mohamed et al. 2020, S. Li et al. 2021), where low-latency inference is critical. Low-latency inference is a capability that should also be applicable to forecasting tasks, which enables efficient real-time predictions in resource-constrained environments.

However, CNNs do come with notable limitations. Their fixed receptive field can hinder long-term forecasting if crucial information lies far in the past, requiring deep stacking or dilated convolutions to extend memory (Chang et al. 2017). Interpretability remains another challenge. While visualizing learned filters can offer insights, they generally provide less transparency than statistical models or attention-based architectures. This issue is further compounded in time series forecasting, where traditional saliency methods, effective in vision tasks, often fail to identify important features over time due to the conflation of temporal and feature dimensions, making it difficult to distinguish meaningful signals at specific time steps (Abdelsalam Ismail et al. 2020). Despite these drawbacks, CNNs shine in speed-critical applications and are often used in hybrid architectures, serving as feature extractors before feeding structured outputs into RNNs or Transformers for long-term dependencies (Livieris et al. 2020, Rostamian and O'Hara 2022, Alshingiti et al. 2023). However, for tasks requiring long memory retention, non-stationary adaptability, or position-sensitive forecasting, alternative architectures may be prefered.

### 4.2.3 Transformers

Originally developed for language tasks, Transformers have gained popularity in time series forecasting due to their ability to capture long-range dependencies through self-attention (Vaswani et al. 2017). Unlike RNNs, which propagate information sequentially, Transformers allow every time step to directly attend to all others, theoretically making them particularly effective for long-horizon forecasting and complex multivariate interactions. This adaptability has led to specialized variants like TFT(Lim, Arik, et al. 2019), Informer (Zhou et al. 2021), Autoformer (Wu et al. 2021), and PatchTST (Nie et al. 2023). However, the interpretability of Transformers in time series remains an open challenge, as self-attention scores do not always align with meaningful temporal dependencies, making it difficult to extract causal relationships from learned representations. While attention maps can offer some insights (Lim, Arik, et al. 2019), their reliance on learned weights rather than explicit time-based reasoning limits their usefulness in high-stakes forecasting applications, such as healthcare or finance.

Empirical studies have demonstrated state-of-the-art accuracy from Transformers in retail demand forecasting, energy load prediction, and financial time series modeling (C. Wang et al. 2022, Bilokon and Qiu 2023, Oliveira and Ramos 2024, Moosbrugger et al. 2025). In retail, Transformer-based models improved forecast accuracy by 26–29% (MASE) over classical methods like ARIMA and ETS, while in energy forecasting, they captured seasonal patterns and demand spikes more effectively than LSTMs (Bilokon and Qiu 2023).

However, Transformers also have inherent limitations. Their permutation-invariant self-attention can lead to temporal information loss, even with positional encodings, making it difficult to fully retain sequential structure(Zeng et al. 2023). Additionally, studies reveal that Transformers often fail to extract meaningful temporal relations from long input sequences, with forecasting errors remaining stable or even increasing as look-back windows grow (Zeng et al. 2023). Surprisingly, a one-layer linear model (LTSF-Linear) outperformed Transformers on multiple benchmarks by 20–50%, raising questions about their true effectiveness in long-term time series forecasting (Zeng et al. 2023).

Another major drawback is computational cost. Standard self-attention scales as $O(L^2)$ in sequence length, making Transformers resource-intensive for long time series. Various efficiency improvements, such as sparse attention patterns (Zhou et al. 2021) and patching (Nie et al. 2023), have been proposed to mitigate this issue. Despite these challenges, Transformers remain a strong choice when the problem complexity justifies the added computational burden, particularly in large-scale forecasting applications where even small accuracy gains translate into significant financial impact (Bilokon and Qiu 2023).

#### 4.2.4 GNNs

Many forecasting problems extend beyond independent time series, especially in supply chain, energy, and traffic systems, where time series function as nodes in a network with edges capturing relationships. Incorporating graph structure into forecasting models can significantly enhance accuracy when interactions among entities matter. For instance, in supply chain forecasting, demand for one product or region may influence another due to substitution effects or shared economic drivers. A study in recent found that graph-based models consistently outperformed traditional statistical and deep learning methods, reducing errors by 10–30% across various supply chain tasks (Wasi et al. 2024).

In energy forecasting, GNNs help model dependencies in power grids, where demand at one substation can influence its neighbors due to load transfer or shared weather conditions (Hu et al. 2022). Similarly, in traffic forecasting (Jiang and Luo 2022), GNN-based models such as Diffusion Convolutional RNNs have demonstrated significant improvements in predicting road speeds by capturing the underlying road network structure. The core advantage of GNNs is their ability to model joint effects across locations, which sequential models ignoring the graph structure would miss.

GNNs introduce additional computational complexity—per time step, the cost is $O(V + E)$, where $V$ is the number of nodes and $E$ is the number of edges. However, many real-world graphs are manageable in size, making GNNs feasible. They are particularly valuable when ignoring the network structure degrades forecasting accuracy, or when the task requires modeling network-specific dependencies. However, if time series are truly independent or lack meaningful structure, a GNN may offer no benefit and could even degrade performance by overfitting spurious connections.

#### 4.2.5 LLMs

LLMs like GPT-3 and GPT-4 have demonstrated exceptional sequence modeling capabilities in natural language processing (NLP), leading researchers to explore their potential for time series forecasting. While LLMs are not inherently designed for numerical time series, their ability to learn general sequence structures has inspired approaches such as tokenizing numeric values or providing structured prompts containing historical data for forecasting. Some methods also leverage LLMs' ability to process multimodal inputs, integrating unstructured data like news and reports to enrich predictions beyond purely statistical patterns (X. Yu et al. 2023).

A key strength of LLMs is their world knowledge (Tang et al. 2025), which enables them to reason about external factors and natural languages that can substantially improve the predictive performance. Unlike traditional forecasting models, which rely solely on past values, LLMs can infer relationships between real-world events and time series dynamics. For instance, an LLM can process news of a company's data breach and forecast a drop in stock prices, performing cross-sequence reasoning that typically requires human interpretation.

However, LLMs face significant challenges in standalone time series forecasting. Studies show that pretraining on text does not translate well to time series, as randomly initialized models often perform as well as pretrained ones. Moreover, LLM-based forecasting models introduce high computational costs without delivering meaningful accuracy improvements, limiting their practicality in real-world, resource-constrained applications (Tan et al. 2024). In TimeGPT's public benchmark, for example, we also observed that model performance deteriorates in high-frequency forecasting (e.g., daily or hourly) compared to lower-frequency settings such as weekly or monthly forecasting (Garza et al. 2023).

Given these limitations, LLMs are currently more effectively used as an augmenting role rather than as standalone forecasters. For example, financial institutions might use traditional time series models for quantitative predictions while employing LLMs to analyze news and adjust forecasts accordingly (Xinlei Wang, Feng, et al. 2024). Meanwhile, foundation models like TimesFM (Das et al. 2024) and Chronos (Ansari et al. 2024) have shown promising results on large-scale benchmarks, hinting at a future where LLMs are specialized for numerical forecasting. Until then, LLMs serve best as complements to traditional models and hard-to-beat baselines, adding domain knowledge, cross-series reasoning, and qualitative insights that purely statistical models cannot.

## 4.3 Emerging Techniques in Forecasting

The period between 2016 and 2025 has seen significant advancements in techniques aimed at enhancing the accuracy, scalability, and robustness of deep learning-based forecasting models. This section presents a chronological survey of key developments and insights on these emerging techniques or tricks that helped the adoption of deep learning models in time series forecasting.

### 4.3.1 Adaptive Input Normalization

Adaptive normalization techniques have been developed to address the challenges of non-stationary time series data. Adaptive Normalization (AdaNorm) (Xu et al. 2019) dynamically adjusts its normalization approach using a learned gating mechanism, selecting between instance normalization, batch normalization, or retaining raw data to improve generalization, particularly for data with shifting distributions. Deep Adaptive Input Normalization (DAIN) (Passalis et al. 2019) employs learned affine transformations to adaptively normalize inputs, enhancing robustness against distributional shifts commonly found in financial and energy forecasting. Reversible Instance Normalization (RevIN) (Kim et al. 2021) applies normalization during training but reverses the transformation at inference, stabilizing forecasts in volatile environments by preserving event-specific patterns. The Non-Stationary Transformer (NSTransformer) (Y. Liu et al. 2022) integrates Series Stationarization to normalize inputs and De-stationary Attention to recover lost non-stationary information, mitigating over-stationarization issues in Transformer-based forecasting models. Unlike DAIN, which utilizes nonlinear transformations for adapting to shifts, and RevIN, which operates at the instance level but sacrifices event-specific features, NSTransformer reconstructs attention weights to enhance generalization, reducing mean squared error (MSE) by up to 49% in long-horizon forecasting (Y. Liu et al. 2022). Most recently, Fourier-based Normalization (FiLM) (Yang et al. 2024) applies frequency filtering to separate low- and high-frequency components before feeding them into the model, reducing noise and emphasizing core temporal patterns, further improving robustness in long-range forecasting tasks.

### 4.3.2 Patching and Residual Learning

The patching method, inspired by vision Transformer architectures, has enhanced forecasting accuracy and efficiency. PatchTST (Nie et al. 2023) segments time series into patches to independently process temporal features, capturing local dependencies and reducing computational overhead. This method demonstrates substantial accuracy improvements for long-horizon, multivariate forecasting tasks. The xPatch introduces a dual-stream forecasting architecture using Exponential Moving Average (EMA) decomposition to separate seasonal and trend components, improving adaptability over Simple Moving Average (SMA)-based smoothing. By integrating patching techniques within a CNN-based framework, xPatch efficiently captures seasonal dependencies while preserving long-term trends, avoiding the permutation-invariance issues of Transformer-based models. This design enhances forecast accuracy, generalization, and computational efficiency, making it more robust than traditional smoothing methods and more scalable than Transformer alternatives (Stitsyuk and Choi 2025). Additionally, residual stacking, exemplified by N-BEATS (Oreshkin et al. 2020), utilizes backward and forward residual blocks to decompose and reconstruct time series patterns without domain-specific knowledge, achieving leading performance in benchmark competitions. Recently, simplified architectures based on MLPs, such as TSMixer (Chen et al. 2023) and TimeMixer (Shiyu Wang, Wu, et al. 2024), have emerged, leveraging channel mixing and temporal mixing strategies to model time series dependencies. Temporal mixing operates by applying MLPs across time steps, enabling the model to capture sequential relationships without the need for self-attention or recurrence. This contrasts with channel mixing, where MLPs process each feature independently across all time steps to learn inter-variable dependencies. By decoupling temporal and feature interactions, these models efficiently learn long-term dependencies while maintaining computational efficiency. Their ability to avoid explicit attention mechanisms significantly reduces computational complexity, allowing them to scale effectively to large-scale forecasting scenarios. This streamlined design enables them to capture complex temporal patterns with remarkable efficiency, making them particularly suited for high-dimensional data where standard architectures may struggle with scalability.

### 4.3.3 Hybrid Decomposition Methods

Hybrid architectures incorporating statistical decomposition emerged prominently with methods like ES-RNN (Smyl 2020) and Autoformer (Wu et al. 2021). ES-RNN combined ETS for trend and seasonality extraction with RNNs for residual modeling, significantly improving accuracy and robustness in diverse forecasting competitions. Autoformer embedded a progressive decomposition architecture within Transformer models, distinctly separating seasonal and trend components, leading to notable accuracy improvements on long-term forecasts.

### 4.3.4 Matrix Factorization Approaches

Matrix factorization techniques have been pivotal in addressing scalability in high-dimensional forecasting contexts. Temporal Regularized Matrix Factorization (TRMF) (Yu et al. 2016) decomposed large-scale time series into latent factors, applying temporal regularizations to capture dependencies effectively, thus providing scalable and efficient training and inference. Sen et al. (2019) further introduced global and local matrix factorization to leverage shared patterns across large collections of series, significantly outperforming traditional approaches in terms of speed and accuracy.

## 5 Case Study: Benchmarking the FTO Framework with M5 Dataset

In this section, we present a comprehensive case study designed to systematically evaluate and compare forecast optimization techniques within the FTO framework, using the established M5 (Makridakis et al. 2022) dataset as a benchmark. The primary objective is to quantify and demonstrate the incremental value of various forecast post-processing methods. To achieve this, we selected a diverse array of deep learning models including NBEATS (Oreshkin et al. 2020), TimesNet (Wu et al. 2022), PatchTST (Nie et al. 2023), TFT (Lim, Arik, et al. 2019), DLinear (Zeng et al. 2023), NLinear (Zeng et al. 2023), NHITS (Challu et al. 2023), KAN (Z. Liu et al. 2024), and TiDE (Das et al. 2023), as well as two prominent foundation models, Chronos-bolt (Ansari et al. 2024) and TimesFM 2.0 (Das et al. 2024). The implementation of the deep time series models were implemented using neuralforecast (Olivares et al. 2022) python package and the two foundation models were implemented using the original GitHub implementation form the authors. The input length of the training data is two times of the forecasting horizon which is 56 days. The architecture-related parameters of each model was kept default. These models were evaluated in conjunction with multiple optimization strategies encompassing BMA (Leamer and Leamer 1978), DLE (NannyML (release 0.13.0) 2023), and TimeSpeaks (Jiang et al. 2025).

BMA was implemented in two distinct variants: BMA Global, wherein a single set of model weights was computed collectively across all SKUs, and BMA Local, which computed individualized model weights tailored specifically for each SKU based on historical forecasting accuracy. The original TimeSpeaks paper demonstrated its implementation using BiLSTM and Transformer architectures. For simplicity, we implemented TimeSpeaks with a fully connected linear network. Additionally, we fitted a separate TimesFM model for DLE for predicting the loss associated with each candidate model. Exogenous variables originally present in the M5 dataset were deliberately excluded from consideration, as the primary aim of this case study is to systematically benchmark and illustrate the efficacy of forecast optimization techniques rather than optimizing predictive performance for the competition itself.

The experimental setup involved generating forecasts across ten sequential windows, each encompassing a 28-day horizon, utilizing the historical training data from the M5 dataset. The dataset comprises 30,490 SKUs spanning 1,941 timestamps. For methods dependent on historical forecasting data, specifically BMA, DLE, and TimeSpeaks, forecasts were simulated across these historical windows to provide the requisite performance insights.

We selected two evaluation metrics to comprehensively assess forecasting performance: Symmetric Mean Absolute Percentage Error (SMAPE), and Weighted Root Mean Squared Scaled Error (WRMSSE). WRMSSE serves as the official evaluation metric for the M5 competition, capturing both scale sensitivity and aggregation consistency across hierarchical levels. All final results reported in this study were evaluated on the M5 competition test set.

Table 3. Comparing FTO methods with other SOTA Deep Learning Forecasting Methods

| Metrics | Forecast then Optimize Methods | | | | Global Deep Learning Models | | | | | | | | | Foundation GenAI Models | |
|---|---|---|---|---|---|---|---|---|---|---|---|---|---|---|---|
| | TimeSpeaks - Linear | DLE - TimesFM | BMA - Global | BMA - Local | Nlinear | Dlinear | NHITS | NBEATS | TimesNet | PatchTST | TFT | KAN | TiDE | TimesFM | Chronos |
| SMAPE | 30.55 | 32.59 | 39.98 | 36.46 | 32.92 | 43.27 | 66.32 | 69.31 | 60.88 | 59.26 | 59.08 | 67.44 | 44.04 | 94.41 | 71.92 |
| WRMSSE | 0.092 | 0.145 | 0.342 | 0.175 | 0.157 | 0.263 | 0.351 | 0.368 | 0.385 | 0.335 | 0.369 | 0.368 | 0.272 | 0.661 | 0.393 |

The results presented in Table 3 reveal several notable insights regarding the comparative performance of post-adjustment techniques and time series forecasting models within the FTO framework. Overall, the findings underscore the importance of forecast post-processing as a critical component in improving predictive performance on structured, hierarchical datasets such as M5.

Among all methods evaluated, TimeSpeaks, implemented with a simple fully connected linear network, achieved the strongest performance across both evaluation metrics, recording the lowest SMAPE of 30.55 and WRMSSE of 0.092. This was followed by DLE, which used TimesFM to predict model loss, with SMAPE of 32.59 and WRMSSE of 0.145. These results demonstrate that post-adjustment techniques, which leverage historical forecasting performance, can significantly outperform raw forecasts generated by advanced deep learning or foundation models. Such findings emphasize that sophisticated post-processing can compensate for individual model weaknesses, effectively synthesizing insights from multiple candidate models.

The two variants of BMA, Global and Local, highlight the utility of personalized ensemble methods. BMA-Local, which tailors model weights per SKU, consistently outperformed BMA-Global across both metrics, reducing SMAPE from 39.98 to 36.46 and WRMSSE from 0.342 to 0.175. This illustrates the advantage of incorporating SKU-level performance data into the ensemble process, as opposed to relying on global model averaging.

Interestingly, foundation models such as TimesFM and Chronos, despite their scale and flexibility, underperformed relative to simpler deep learning models. TimesFM recorded an SMAPE of 94.41 and WRMSSE of 0.661, while Chronos reached SMAPE 71.92 and WRMSSE 0.393. These outcomes suggest that, when exogenous variables are excluded and no fine-tuning is applied, large pretrained models may struggle to generalize effectively for long horizon forecasting. This underperformance can be attributed to several factors. Firstly, these models often employ autoregressive decoding strategies, where each predicted value is fed back into the model to predict subsequent values. This approach can lead to error accumulation over extended forecasting horizons, compounding inaccuracies as predictions progress further into the future. Secondly, without fine-tuning or incorporation of auxiliary data, the generalization capabilities of large pre-trained models diminish in long-horizon forecasting scenarios.

Among the base deep learning models, NLinear achieved a standout performance with SMAPE of 32.92 and WRMSSE of 0.157, outperforming not only more complex architectures such as NBEATS, TFT, and NHITS, but also some of the post-adjustment techniques. For instance, BMA-Global and Chronos yielded notably worse results. This highlights an important insight: the effectiveness of post-processing techniques is conditional on the relative quality and diversity of the candidate models. When a single base model, such as NLinear, already demonstrates strong predictive capability, the marginal gains from ensemble-based post-processing diminish. Conversely, the value of forecast optimization becomes more pronounced when the performance of individual models is more balanced, allowing the optimization framework to capitalize on complementary model strengths.

In summary, the findings from Table 3 highlight the substantial potential of forecast optimization techniques, particularly those that integrate model performance history into adaptive post-processing.

However, the benefits derived from such techniques are context-dependent, with their relative advantage increasing as the performance gap between candidate models narrows. This case study thus offers practical guidance for deploying FTO strategies effectively in production environments.

## 6 Discussion and Future Research Directions

### 6.1 FTO in Operations Management Research

The forecast-then-optimize methodology represents a prevalent paradigm for tackling decision-making problems under uncertainty in operations management. This approach typically involves a two-stage process: first, uncertain parameters that influence the decision problem, such as future demand, prices, or resource availability, are forecasted using various statistical or machine learning techniques. Second, these forecasts are then fed into an optimization model to determine the optimal decisions. This sequential approach allows for the decomposition of complex problems into more manageable steps.

The forecast-then-optimize methodology has widespread application across various domains within operations management. In inventory management and control, forecasting demand is crucial for determining optimal inventory levels, setting reorder points, and minimizing the risk of stockouts or excessive holding costs. In supply chain management, this methodology can be used to forecast demand at different stages of the chain, helping to mitigate the bullwhip effect and improve overall supply chain efficiency. Revenue management relies heavily on forecasting demand to optimize pricing strategies, manage capacity effectively, and maximize revenue. Healthcare operations management utilizes forecast-then-optimize for tasks such as forecasting patient arrivals to optimize staffing, allocate resources efficiently, and improve patient flow. FTO can benefit energy sector applications like renewable energy trading and grid management, where forecasting renewable energy generation is essential for optimizing grid operations and making informed trading decisions. In transportation and logistics, FTO may be used to predict demand for transportation services, enabling the optimization of vehicle routes, delivery schedules, and airline schedules. Even in humanitarian logistics, the framework has potential applications such as optimizing the locations of resources based on forecasted needs.

The advancements in machine learning and deep learning have significantly enhanced the capabilities of the FTO methodology. Machine learning models are increasingly employed in the forecasting stage due to their ability to learn complex patterns and relationships from large datasets, leading to more accurate predictions of uncertain parameters. Deep learning techniques, with their capacity to handle high-dimensional and non-linear data, are being explored for tackling complex forecasting tasks within this framework, particularly for time series data that exhibit intricate patterns.

### 6.2 Summary of Current Gaps and Limitations

Despite advances in deep learning and post-processing, key gaps remain in the FTO framework. First, the trade-off between accuracy and interpretability persists, especially in high-capacity models and sophisticated ensembles where forecast adjustments can obscure the reasoning behind predictions. In critical decision-making domains, such as healthcare, finance, and supply chain, stakeholders require transparent reasoning behind forecasts, yet many FTO-driven models do not readily provide explanations. Recent studies underline that improving interpretability is crucial for trustworthiness and adoption of forecasting tools (Kim et al. 2025). Yet the reliability of post-hoc explanation techniques in time series contexts is not fully proven and the complexity tends to increase (Turbé et al. 2023). This lack of interpretability can be a serious barrier in high-stakes settings where model outputs affect policy or significant business decisions. Second, although techniques like meta-learning and dynamic ensembles have improved accuracy, their generalization and robustness under distribution shifts or across heterogeneous time series remain limited. Finally, most FTO workflows assume batch processing, while real-time forecasting and online adjustment are still underdeveloped due to challenges in latency, data drift, and the cost of dynamic refinement during inference. Bridging these gaps is critical for bringing FTO into real-world, adaptive decision-making environments.

### 6.3   Recommendations for Future Research

To advance the FTO paradigm, future research should focus on several key areas. First, the development of lightweight, distributed frameworks for residual learning and dynamic model selection will be crucial for enabling real-time adaptability without incurring significant computational overhead. These systems should be designed with scalability in mind, allowing seamless integration into online forecasting pipelines with minimal latency.

Second, enhancing transferability of meta-learning strategies and ensemble methods remains a vital challenge. Research should explore techniques that promote domain adaption and robustness across heterogeneous time series with minimal fine-tuning.

Third, Human-in-the-Loop (HITL) FTO systems represent an underexplored but high-impact direction. Incorporating expert feedback, override mechanisms, or interactive post-processing layers—such as counterfactual explanations or scenario-based refinement—can significantly improve model interpretability and stakeholder trust, particularly in high-stakes or regulated environments. Looking further ahead, another exciting recommendation is to develop agent-based forecasting systems that learn to forecast and act in a rapid environment. A recent development is the use of LLM-based agents for event forecasting, where a language model with tools and memory autonomously gathers information and generates forecasts for geopolitical events (Ye et al. 2024). We recommend exploratory work in this direction: creating autonomous forecasting agents that can adapt, learn, and possibly collaborate (Halawi et al. 2024). For instance, an agent could use online learning to continuously refine its model parameters as new data comes, or multiple agents could negotiate a consensus forecast.

# 7 Conclusion

Time series forecasting has experienced a significant evolution over the past decade, driven by the rise of deep learning architectures that offer increased flexibility, scalability, and domain adaptability. As forecasting systems become more embedded in high-stakes decision-making, the need for robust and interpretable post-model refinement has become increasingly clear. The FTO framework addresses this challenge by formalizing a structured, model-agnostic approach to enhance forecasts through downstream adjustments.

This review has outlined the core components of the FTO framework, surveyed state-of-the-art deep learning models, and analyzed a broad spectrum of post-processing techniques—including ensembles, meta-learners, and uncertainty-aware refinements. Together, these advancements signal a paradigm shift: from model-centric innovation to outcome-centric refinement.

As forecasting moves beyond academic benchmarks into operational, real-world environments, the importance of reliable, interpretable, and adaptive post-processing cannot be overstated. Reliable, interpretable, and adaptive FTO pipelines can bridge the gap between prediction and decision-making, empowering organizations to respond to uncertainty with speed and confidence. Future research in FTO holds the potential to unify prediction, optimization, and uncertainty into a cohesive pipeline—offering a promising path toward more intelligent and autonomous decision systems.